\documentclass[sigconf]{acmart}

\AtBeginDocument{%
  \providecommand\BibTeX{{%
    \normalfont B\kern-0.5em{\scshape i\kern-0.25em b}\kern-0.8em\TeX}}}

\copyrightyear{2020}
\acmYear{2020}
\setcopyright{acmlicensed}
\acmConference[ICMR '20]{Proceedings of the 2020 International Conference on Multimedia Retrieval}{June 8--11, 2020}{Dublin, Ireland}
\acmBooktitle{Proceedings of the 2020 International Conference on Multimedia Retrieval (ICMR '20), June 8--11, 2020, Dublin, Ireland}
\acmPrice{15.00}
\acmDOI{10.1145/3372278.3390736}
\acmISBN{978-1-4503-7087-5/20/06}

\usepackage{mathtools, nccmath}
\usepackage{titlesec}
\usepackage{balance} 

\begin{document}
  

\fancyhead{}

\title{Salienteye: Maximizing Engagement While Maintaining Artistic Style on Instagram Using Deep Neural Networks}

\author{Lili Wang}
\email{lili.wang.gr@dartmouth.edu}
\affiliation{%
  \institution{Dartmouth College}
  \city{Hanover}
  \state{New Hampshire}
  \country{USA}
}
\author{Ruibo Liu}
\email{ruibo.liu.gr@dartmouth.edu}
\affiliation{%
  \institution{Dartmouth College}
  \city{Hanover}
  \state{New Hampshire}
  \country{USA}
}
\author{Soroush Vosoughi}
\email{soroush.vosoughi@dartmouth.edu}
\affiliation{%
  \institution{Dartmouth College}
  \city{Hanover}
  \state{New Hampshire}
  \country{USA}
}

\begin{abstract}

Instagram has become a great venue for amateur and professional photographers alike to showcase their work. It has, in other words, democratized photography. Generally, photographers take thousands of photos in a session, from which they pick a few to showcase their work on Instagram. Photographers trying to build a reputation on Instagram have to strike a balance between maximizing their followers' engagement with their photos, while also maintaining their artistic style. We used transfer learning to adapt Xception, which is a model for object recognition trained on the ImageNet dataset, to the task of engagement prediction and utilized Gram matrices generated from VGG19, another object recognition model trained on ImageNet, for the task of style similarity measurement on photos posted on Instagram. Our models can be trained on individual Instagram accounts to create personalized engagement prediction and style similarity models. Once trained on their accounts, users can have new photos sorted based on predicted engagement and style similarity to their previous work, thus enabling them to upload photos that not only have the potential to maximize engagement from their followers but also maintain their style of photography. We trained and validated our models on several Instagram accounts, showing it to be adept at both tasks, also outperforming several baseline models and human annotators.
 
\end{abstract}


\maketitle

\section{Introduction}
Amongst the top social media platforms (such as Facebook, Instagram, and Twitter) \cite{perrin2019nd}, Instagram has established itself as one of the main communities for amateur and professional photographers to share their work. Most artists trying to build a reputation often have to deal with two, and sometimes competing, priorities: profit and artistic integrity. The same is true for photographers trying to build a following on Instagram, except that on Instagram, the currency of profit is engagement. Using the number of likes on a photo as a proxy for engagement and the style of a photo (e.g., closeups, use of filters, etc) as a proxy for a photographer's artistic sensibilities, we created a tool called SalientEye that once trained on any individual Instagram account, it can sift through new photos by the same user and sort them based on predicted future engagement and proximity to the user's style.  

SalientEye is comprised of two models, one for predicting engagement and one for measuring style similarity. Typically, to get high-performing models for such tasks, a deep neural network would be trained using millions of photos. However, we used Xception \cite{Xception} and VGG19 \cite{VGG}, models pre-trained on ImageNet \cite{deng2009imagenet} (a large dataset of labelled high resolution images with around 22,000 categories) for our task. We used transfer learning, which enables us to take a pre-trained model and fine-tune it to a new (but related) task with only a few thousand data points \cite{transfer}, to create an engagement prediction model from Xception for photos posted on Instagram. We also used Gram matrices generated from VGG19 for the task of style similarity measurement on photos posted on Instagram. 

Thus, we were able to create engagement prediction and style similarity models for Instagram without a need for a massive dataset or expensive training. These models can be trained on any given Instagram account, creating a personalized engagement prediction and style similarity model for that account. After the models have been trained, they help users decide which photos to post on their accounts from a new batch of photos by sorting the photos according to their predicted engagement and similarity to the user's style. The importance of personalization for the style similarity model is self-evident, as each user might have vastly different styles. However, it is also important that the engagement prediction model is personalized. The followers of one account might prefer different types of photos to the followers of another account. So even if the number of followers of two accounts is the same, it is very possible that a photo would generate different amounts of engagement in those accounts. In other words, the personalized engagement prediction model learns the "taste" of an account's followers.

\section{Related Work}
As image sharing communities have grown over the last few years, there has been an increasing number of works attempting to study and predict the engagement of users with photos on these communities, particularly on Flickr and Instagram. Zhu et al. \cite{filter}, show that photos using filters are more likely to be viewed and
commented on, while Bakhshi et al. \cite{faces}, show that photos with faces attract more likes and comments. Several works have attempted to predict engagement using various combinations of text-based, network-based, and user-based features \cite{niu2012predicting,mcparlane2014nobody,khosla2014makes,totti2014impact,deza2015understanding,zhang2018become,zhang2018user,hessel2017cats,ding2019intrinsic}. However, all of these works use a large dataset, containing many users. Though this allows for training generalized models of engagement, they do not capture the differences in engagement dynamics between different accounts (since engagement is a function of the followers of an account, one can expect the dynamics of engagement to be different for each account, based on the type of followers an account has). Mazloom et al. \cite{mazloom2016multimodal,mazloom2018category} have more granular models of engagement, looking at category-specific posts. The work that gets closest to having truly user-specific models of engagement is the work of Zhang et al, \cite{zhang2018become}, however, they also train their models on a dataset from a few hundred users. 

Work on texture style modeling can be traced back to a paper by Julesz in 1962 \cite{julesz1962visual}. In this paper, Julesz, showed that a visual texture can be uniquely described by the Nth-order statistics. Since the late 1990s, there has been several influential papers on texture synthesis modelling, using complex wavelet coefficients \cite{portilla2000parametric}, Markov random field \cite{efros1999texture}, tree-structured vector quantization \cite{wei2000fast}, and combination of filtering theory and Markov random fields \cite{zhu1998filters}. More recently, in 2015, Gatys et al. \cite{gatys2015texture} and Lin et al. \cite{lin2015bilinear} showed that the Gram matrix representations extracted from the VGG19 object recognition model \cite{VGG}, can model the textural style well. Our style analysis is based on these works.

To the best of our knowledge, our model is the first user-specific model that can be trained on any single Instagram account, relying purely on image-based features. Besides predicting future engagement with new photos of a photographer, we also rank the new photos based on style similarity to the photographer's previous work, allowing them to maintain their style while maximizing engagement with their work. 

\section{Methods}

\subsection{Definition of engagement on Instagram}

Previous work on analyzing engagement on Instagram \cite{niu2012predicting,mcparlane2014nobody,khosla2014makes,totti2014impact,deza2015understanding,zhang2018become,zhang2018user} treated images posted on different time periods the same way. However, as an account gets older and gathers more followers, the average number of likes for new photos posted goes up. So everything else being equal, the same photo posted when an account is younger will have less likes now than if it was posted when the account was older. 
Thus, on average, photos posted later in the life of an account tend to get more likes.

Based on this observation, in order to determine the engagement of a photo we looked through all the photos posted by the same account one month before and one month after: if the number of likes of the photo was in the bottom third (i.e., below the first tertile), we considered the photo to be unpopular (i.e., have low engagement); if the number of likes was in the top third (above the second tertile), we considered it to be a popular (high engagement) photo. If the photo fell between the first and second tertiles, we considered the photo to have average engagement. Through this method, we are defining high and low engagement in the context of average engagement of photos posted on the same account around the same time. The assumption here is that the number of followers and the taste of the followers of an account is relatively stable in a short (two months) time frame.

\subsection{Data}

For this paper, we selected 21 Instagram accounts,. The photos posted on these accounts are quite diverse, featuring nature, people, architecture, landscapes and animals (both wild and domestic). We collected all the images from these accounts. We used all 21 accounts for fine-tuning our models' parameters (explained in the next section) and used seven of these accounts in our experiments, (Table \ref{tab:21large} shows the total number of posts and followers for each of these seven accounts).  The seven accounts were chosen to get a mixture of amateur and professional photographers. Also, four of the seven accounts are related to National Geographic (NatGeo), meaning that they have very similar styles, while the other three are completely unrelated. This allows us to better test our style similarity models (this is explained in more detail in the section on experiments). We used the method explained in the last section to create a dataset of photos with high and low engagement for each account.

\subsection{Engagement prediction}
Using the dataset described above, we utilized transfer learning to fine-tune several off-the-shelf, pre-trained models to our Instagram dataset. For each pre-trained model, we first fine-tuned the parameters using the photos in our dataset (from the 21 accounts), dividing them into a training set of 23,860 images and a validation set of 8,211. We only used photos posted before 2018 for fine-tuning the parameters since our experiments (discussed later in the paper) used photos posted after 2018. Note that these parameters are not fine-tuned to a specific account but to all the accounts (you can think of this as tuning the parameters of the models to Instagram photos in general).

Using the validation set, we fine-tuned and evaluated several state-of-the-art, pre-trained models; specifically, we looked at \emph{VGG19} \cite{VGG}, \emph{ResNet50} \cite{Resnet}, \emph{Xception} \cite{Xception}, \emph{InceptionV3} \cite{Inception} and \emph{MobileNetV2} \cite{Mobilenet}. All of these are object recognition models pre-trained on \emph{ImageNet}\cite{deng2009imagenet}, which is a large dataset for object recognition task. We picked Xception as it had the highest validation performance ($F1=0.69$).

To fine-tune the models, we used a common method, developed in recent years, for deep transfer learning for computer vision models \cite{transfer}. We modified the object recognition models to serve as engagement prediction models using standard methods. Here we describe how this was done for Xception (which is the model we ended up using): we froze the first 60 layers of Xception and replaced the ImageNet top layer with one global average pooling layer and two fully connected layers. The first fully connected layer has 1024 neurons with 'ReLU' activation, and the second layer has two output nodes (high or low engagement) with 'softmax' activation. We initialized the learning rate with 0.005 and used stochastic gradient descent (SGD) optimizer with a momentum of 0.9 and a decay of 1e-6 trained for a total of 10 epochs with a batch size of 64. 

It is important to note that the modified Xception model at this stage has its parameters fine-tuned on several Instagram accounts. As we show later in the paper when we discuss the experiments, this model can now be trained on individual accounts to create account-specific engagement prediction models.

Also, note that our engagement prediction model is based purely on the image. We do not consider text and hashtags in predicting engagement (though they likely have some predictive power) as the point of our tool is to sort through large photo albums and make recommendations for photos to be posted on Instagram. 

\subsection{Style similarity}
Motivated by Gatys et al.'s work on style synthesis (more specifically, texture synthesis) \cite{gatys2015texture}, we used Gram matrices for our style similarity model. Gatys et al. \cite{gatys2015texture}, and Lin et al. \cite{lin2015bilinear} showed that Gram matrix representations extracted from each layers of \emph{VGG19} \cite{VGG} can model the textural style well. For instance, Gatys et al., \cite{gatys2015texture} used the difference of Gram matrices of a white-noise image and a texture image (e.g., rocks) as a part of the loss function to synthesize a new texture image. Different from what Gatys et al. do, we apply the Gram matrix method to measure the style similarity of two non-texture images.

Our style similarity model works as follow. Assume $F_{jk}^l$ is the activation of the $j^{th}$ filter at position k in layer l of an image from the \emph{VGG-19} network. The Gram matrix of the image is defined as:

\useshortskip
\begin{equation}
G_{ij}^l =\sum_{k} F_{ik}^lF_{jk}^l
\end{equation}

Assume a and b are the images that we are comparing the style of, and $G_{a}^l$ and $G_{b}^l$ are their Gram matrix representations in layer l. The total style difference is defined as:
\useshortskip
\begin{equation}
F(a,b)= \sum_{l=0}^{L} w_{l}\frac{1}{4N_{l}^2M_{l}^2} \sum_{i,j}(G_{a ij}^l-G_{b ij}^l)^2
\end{equation}
Where $N_{l}$ is the number of filters in layer l, $M_{l}$ is the size of each filter in layer l.

For style synthesis, all the $w_{l}$ are set to the same value. In our work, however, we are interested in capturing high-level and more abstract style instead of low-level texture and patterns, thus we set $w_{l}$ to be larger for deeper layers (since the deeper layers capture more abstract style).

\begin{table*}[ht]

\footnotesize
\caption{Confusion matrix for the task of account prediction using style similarity. The performances of SalientEye (S) and human (H) annotators are shown.}
  \label{tab:c7}
\begin{tabular}{|l|c|c|c|c|c|c|c|c|c|c|c|c|c|c|}
\toprule

  & \multicolumn{2}{l|}{natgeo (NG)}& \multicolumn{2}{l|}{natgeomagarab (NGM)}& \multicolumn{2}{l|}{natgeotravel (NGT)}& \multicolumn{2}{l|}{thephotosociety (TPS)} & \multicolumn{2}{l|}{cats{\_}of{\_}instagram (COI)} & \multicolumn{2}{l|}{travelalberta (TA)} & \multicolumn{2}{l|}{clarklittle (CL)} \\ 
                   
                    \midrule
                    & S            & H            & S                   & H                  & S                & H               & S               & H    & S               & H   & S               & H   & S               & H             \\ 
                    \midrule
NG     &  \textbf{21\%}  &  20\%    &  12\%  &  26.67\%  & 13\%  &  13.33\%  &  14\% & 16.67\% & 13\%  &  13.33\% & 15\%  & 6.67\%  &  12\%  &    3.33\%  \\ 
NGM  &   14\%  & 23.33\%   &  \textbf{23\%} & 20\%  & 7\%  & 13.33\%   & 18\%   & 26.67\% & 10\%  & 3.33\%  &  21\%  & 13.33\%  &  7\%  &  0\%    \\ 
NGT    &    7\% &   10\% & 9\%    & 26.67\%  & 14\%  &  \textbf{23.33\%}  & 20\%  &  3.33\%   &   5\% & 16.67\%  & 25\%  & 16.67\%  & 20\%  &   3.33\%      \\
TPS  &  15\%  &  23.33\%   &  16\% & 26.67\%   & 11\%  & 13.33\%  &  \textbf{24\%}  & 13.33\% & 15\%  &  16.67\%  & 14\%  &  0\%  &  5\% &   6.67\%   \\ 
COI &   3\% &   10\%  &  3\%  &  30\%   &  1\% & 6.67\%   &   5\%  & 3.33\%&   \textbf{77\%} &  40\% &  9\% & 3.33\%  & 2\%   & 6.67\%     \\ 
TA  &  6\%  &  3.33\%   & 8\%  &  10\%  &  17\%  &  36.67\%  &   7\%  & 13.33\% &  5\% &  6.67\% & \textbf{49\%}  & 26.67\%  &  8\% &    3.33\%   \\ 
CL  &   4\% &  16.67\%   &  2\%  & 10\%   & 18\%  &  16.67\%  &  8\%  &  16.67\%&  1\%  & 10\%  & 9\%  & 3.33\%  & \textbf{58\%}  &26.67\%      \\ 
    \bottomrule
\end{tabular}
\end{table*}

\section{Experiments}
\subsection{Engagement prediction}  
For each of the seven accounts shown in Table \ref{tab:21large}, we used the transfer learning framework explained earlier \cite{transfer} to train our engagement model on all the photos from that account, with the exception of the photos posted in the year 2018, as those were kept for testing. This generated seven user-specific engagement prediction models which were evaluated on the test dataset for each account.

Since there is no previous research on user-specific engagement prediction, we compared our model against three closely related models and human annotators: Hessel2017\cite{hessel2017cats} and Ding2019\cite{ding2019intrinsic} are two state-of-the-art models, LikelyAI is a commercialized product which can be found at \url{https://www.likelyai.com/}. These two state-of-the-art models are trained on a large mixed dataset to predict the popularity score of an image. In order to directly compare these models against ours, we mapped the popularity scores generated by these models into high/low engagement labels. To make sure that our choice of threshold does not negatively affect the performance of these models, we tried all possible binning of their scores into high/low engagement and picked the one that resulted in the best F1 score for the models we are comparing against (on our test dataset). Moreover, we tested both the pre-trained models (which the authors have made available) and the models trained on our dataset and report the best one. Through these two steps (picking the best threshold and model) we can be confident that our comparison is fair and does not artificially lower the other models' performance. 

We used  Amazon Turk to measure the capability of human annotators on predicting the engagement of a photo. For each of the seven accounts, we created a photo album with all the photos that were used to train our models. The photos were divided into high and low engagement. We had three annotators independently examine the photo albums. The annotators were then shown 10 photos randomly selected from our test-set (5 with high engagement and 5 with low) and asked them to predict whether these photos will have high or low engagement. Since each photo was labelled by three annotators, we used the majority label as the label (Fleiss' $\kappa$ = 0.21). The same photos that were used for the human annotation task (10 photos per Instagram account) were also passed to the LikelyAI API (LikelyAI is a commercial product and they do not make their model available for replication).

The performance of these models and the human annotators can be seen in Table \ref{tab:21large}. We report the macro F1 scores of these models and the human annotators. Our model outperformed the state-of-the-art models and human annotators in all the experiments except for the \emph{natgeo} account, where human annotators got a perfect majority score. Upon further examination we see that almost all the high engagement photos in this account are of animals and low engagement photos are of people, which made it an easy task for human annotators. Whenever there is such a clear separation of categories for high and low engagement photos, we can expect humans to outperform our models.

Our experiment also showed that in general the engagement of photos with people is harder to predict. We speculate that this might be because photos with people have a much higher variance when it comes to engagement (for instance pictures of celebrities generally have very high engagement while pictures of random people have very little engagement). Our experiment also showed that the engagement of pictures of nature is easier to predict. In general, the performance of each user-specific model is dependent on the number of photos posted and the number of followers: the accounts with more posts have more training data and the accounts with more followers most likely have more accurate labels (i.e., more clear separation between high and low engagement photos).

\begin{table}[hb]

\footnotesize
  \caption{Performance (macro F1 score) of engagement prediction of SalientEye (S) Hessel2017 (H) Ding2019 (D) LikelyAI (L) and human (M) annotators.}
  \label{tab:21large}
  \begin{tabular}{|l|l|l|l|l|l|l|l|}
    \toprule
    &Posts &Followers & F1{\_}D & F1{\_}S & F1{\_}H & F1{\_}M & F1{\_}L\\
    \midrule
    NG& 18465 &121M &  0.6453 & 0.8727  &0.4631 &\textbf{1.0}  &0.4000\\
    TA & 3344 &672K & 0.7641 & \textbf{0.8266}  &0.5797 &0.4950 &0.6969\\
    TPS & 7948 &5.1M& 0.6446 & \textbf{0.8244} &0.4721 &0.7917  &0.4505\\
    COI & 7040 & 10.6M& 0.7028 & \textbf{0.7195} &0.5962 &0.1667 &0.6000\\
    NGT & 9995 &34M & 0.6752 & \textbf{0.7184}  & 0.5100 &0.4949 &0.5833\\
    NGM & 9580 &540K&  0.7231 & \textbf{0.7181} & 0.6114 &0.4949  & 0.6000\\
    CL& 5305 &2.2M &  0.6838 & \textbf{0.7148}  &0.5359 &0.3333  &0.5833\\

    \bottomrule
  \end{tabular}
\end{table}

\subsection{Style similarity}
To evaluate the performance of our style similarity model, we devised an experiment where we used the style similarity model to predict which account a photo came from, using only their style. We selected the 200 most recent photos from each of the seven accounts. Of those photos, the 100 most recent photos are used to create a test set and the other 100 photos are used as reference photos to represent the style of the account. We setup the experiment in this way to capture the most recent style of an account and judge the style similarity of "new" photos coming in. For each photo in the test set, we used Equation 2 to calculate the style similarity between the photo and each of the 100 reference photos. Since an account could have several different styles, we add the top 30 (out of 100) similarity scores to generate a total style similarity score. We then assign the account with the highest similarity score to be predicted origin account of the test photo.

\begin{table}

\fontsize{6.3}{8}\selectfont

\caption{Confusion matrix for account prediction using style similarity.  Here the NatGeo related accounts are all combined into one set, called NC. The performances of SalientEye (S) and human (H) annotators are shown.}
  \label{tab:natgeo}
\begin{tabular}{|c|c|c|c|c|c|c|c|c|}
\toprule
                    & \multicolumn{2}{c|}{NC} & \multicolumn{2}{c|}{COI} & \multicolumn{2}{c|}{TA} & \multicolumn{2}{c|}{CL} \\ 
                    \midrule
                    & S            & H            & S                   & H                  & S                & H               & S               & H              \\ 
                    \midrule
NC            & \textbf{53.5\%}    & 44.17\%         &  13\%     &  22.5\%    &  22.25\%      & 22.5\%     & 11.25\%   & 10.84\% \\ 
COI & 7.25\%   &  9.17\%    &    \textbf{80.25\%}         &  74.17\%     &  9.75\%                &      11.66\%          &        2.75\%         &      5\%   \\ 
TA       &    31\%   &   25.83\% &   8.5\%        &     8.33\%       &  52.5\%                &  \textbf{53.33\%}  &    8\%    &   12.5\%   \\ 
CL         &    25.75\%    & 29.17\%  &   3\%        &   12.5\%     &    11.75\%            &    13.34\%         &   \textbf{59.5\%}    &   45\%    \\ 
    \bottomrule
\end{tabular}
\end{table}

For comparison, we again used Amazon Turk to measure the capability of human annotators to comprehend the style of an account. For each of the seven accounts, we created a photo album with the 100 reference photos used by our model above. We asked the annotators to pay close attention to the style of each account. The annotators were shown 10 photos randomly selected from the test set of 100 photos of each of the seven accounts (so a total of 70 photos, shown in random order). We then asked the annotators to guess which album the photos belong to based only on the style. Table \ref{tab:c7} shows the confusion matrix for both SalientEye and the human annotators. Our model outperformed the human annotators on all accounts, except for \emph{natgeotravel}.  

Note that there was high confusion between the top four accounts for both the model and the human annotators. These are all accounts that post photos by NatGeo photographers, meaning that the style of the photos posted are very similar in these accounts (\emph{thephotosociety} is a collective of NatGeo photographers). The photos from the other 3 accounts are much better predicted by our model (less so by the humans). To make this point more clear, we repeated the experiments, this time combining the photos from the four NatGeo related accounts into one account (called \emph{natgeo\_collection (NC)}). Table \ref{tab:natgeo} shows the confusion matrix from this experiment. Here, there is much less confusion between the accounts for both our model and the annotators (Fleiss' $\kappa$ = 0.19). 

It should be noted that for both the style and engagement experiments we created anonymous photo albums without any links or clues as to where the photos came from.

\section{Conclusions}
In this paper, we introduced SalientEye, a tool designed to help users select the best (based on the two criteria mentioned above) photos to post on their Instagram accounts. We used transfer learning to adapt Xception which is a model for object recognition to the task of engagement prediction and utilized Gram matrices generated from VGG19, for the task of style similarity measurement on Instagram. 

SalientEye can be trained on individual Instagram accounts, needing only several hundred photos for an account. We tested SalientEye on seven accounts, comprising of both amateur and professional photographers, showing that on average, it is adapt at predicting both the level of engagement of a new photo and its style similarity to a user's previous photos, also outperforming all the other state-of-the-art models and human annotators in both tasks. Instagram has already started to democratize photography, SalientEye has the potential to speed up the process by making it easier for photographers to create the optimal portfolio on Instagram.

To facilitate future research, the implementation of our tool is available upon request.

\bibliographystyle{ACM-Reference-Format}
\balance 
\bibliography{main}

\end{document}